\documentclass[sigconf]{acmart} 

\usepackage[super]{nth}
\usepackage{booktabs} 

\usepackage{draftwatermark}
\SetWatermarkText{Draft}
\SetWatermarkScale{1}
\SetWatermarkLightness{0.8}

\setcopyright{acmlicensed}
\renewcommand\footnotetextcopyrightpermission[1]{} 
\acmConference[]{}{}{}
\acmPrice{}

\begin{document}
\title{Kappa Learning: A New Method for Measuring Similarity Between Educational Items Using Performance Data}

\author{Tanya Nazaretsky}
\affiliation{%
  \institution{Weizmann Institute of Science}
}
\email{tanya.nazaretsky@weizmann.ac.il}

\author{Sara Hershkovitz}
\affiliation{%
  \institution{The Center for Educational Technology}
}
\email{sarah@cet.ac.il}

\author{Giora Alexandron}
\affiliation{%
  \institution{Weizmann Institute of Science}
}
\email{giora.alexandron@weizmann.ac.il}

\renewcommand{\shortauthors}{T. Nazaretsky et al.}

\begin{abstract}
Sequencing items in adaptive learning systems typically relies on a large pool of interactive assessment items (questions) that are analyzed into a hierarchy of skills or Knowledge Components (KCs). Educational data mining techniques can be used to analyze students performance data in order to optimize the mapping of items to KCs. 
Standard methods that map items into KCs using item-similarity measures make the implicit assumption that students' performance on  items that depend on the same skill should be similar. This assumption holds if the latent trait (mastery of the underlying skill) is relatively fixed during students' activity, as in the context of \textit{testing}, which is the primary context in which these measures were developed and applied. However, in adaptive learning systems that aim for \textit{learning}, and address subject matters such as K6 Math that consist of multiple sub-skills (e.g., ``adding fractions with common denominator''), this assumption \textit{does not} hold. 
In this paper we propose a new item-similarity measure, termed \textit{Kappa Learning} (KL), which aims to address this gap.  KL identifies similarity between items under the assumption of \textit{learning}, namely, that learners' mastery of the underlying skills changes as they progress through the items.
We evaluate Kappa Learning on data from a computerized tutor that teaches Fractions for \nth{4} grade, with experts' tagging as ground truth, and on simulated data. Our results show that clustering that is based on Kappa Learning outperforms clustering that is based on commonly used similarity measures (Cohen's Kappa, Yule, and Pearson). 
\end{abstract}

%
%



\maketitle

\section{Introduction}
\label{intro}
Mastery learning \cite{bloom68} is based on the assumption that the domain knowledge can be analyzed into a hierarchy of component skills, with prerequisites between them \cite{corbett1994knowledge,corbett_stuent_modeling_mastery}. This structure can be used to sequence learning in an Intelligent Tutoring Systems so that students master prerequisite skills before moving to skills that depend upon them \cite{corbett1994knowledge}. \textit{Cognitive model} is a formal representation of this structure that is encoded into the tutor. It is typically generated in a process that relies on domain experts, learning scientists, and programmers \cite{cen_model_improvement}. 

A significant part of this process is the mapping of question items into the skills that underlie them (skills are also referred to as Knowledge Components, abbreviated KCs\footnote{in the Psychometrics literature, skills are also referred to as \textit{latent factors} or \textit{constructs}}; in this paper we use the two terms interchangeably). Q-matrix is a standard representation used in Psychometrics to specify the relationships between individual test items and target skills \cite{tatsuoka_qmatrix}. Generating item-to-skill mapping requires a significant human-labor and expertise \cite{Desmarais_bayes_students_models}. In addition, evidence shows that experts' mapping of items into skills can be significantly inconsistent with students' learning process \cite{sunbok_efa}. Thus, methods that identify the skills underlying each item, or assist human experts in doing so, can optimize the process by increasing its accuracy and reducing human labor \cite{Desmarais_qmatrix_enhancing,Koedinger_model_improvement}. 
 
Constructing Q-Matrix from response data is an active
research topic. Barnes \cite{barnes2005q} ``mined'' students' data to create concept models that can be used to direct learning paths. Examples within the Psychometrics literature include \cite{tatsuoka_qmatrix, De_La_Torre,Data-Driven-Learning-of-Q-Matrix}. 
A Matrix Factorization-based method for Q-matrix construction was proposed in \cite{Desmarais_qmatrix}, and was later used for enhancing expert-based Q-Matrices \cite{Desmarais_qmatrix_enhancing}. 
Learning Factor Analysis (LFA) \cite{cen_model_improvement} is a combinatorial search algorithm for optimizing the cognitive model while controlling for model complexity. In \cite{liu_closing} it was demonstrated that using LFA to refine the human-generated cognitive model of a tutor improves learning gains. Performance Factor Analysis (PFA) \cite{Pavlik:2009:PFA:1659450.1659529} reconfigured LFA to enable predictions for individual students with individual skills (LFA assumes all students accumulate learning at the same rate), and also addresses the multiple KCs problem (standard Knowledge Tracing \cite{corbett1994knowledge} assumes that each item requires one KC; examples of extensions that address multiple skills include \cite{peter_subskills,mostow_subskills}). A different approach for `human-in-the-loop' Student Model Discovery (finding the item-to-skill assignment that best describe students' behavior) was proposed in \cite{stamper_human_machine}. 
 

In general, there are two approaches for mapping items into skills: \textit{Model-based}, and \textit{Similarity-based} \cite{Rihak2017}. Model-based methods reduce the dimensionality of the problem and try to infer the latent factors (=skills or KCs) that underlie the items. The methods mentioned above fall into this category. Similarity-based approaches are based on the assumption that students will tend to have similar performance on items that require the same skill, thus seek to identify the similarity between pairs of items. Examples of methods that are based on item-similarity measures include \cite{Rihak2017,trevor_similarity}.


The first phase of item similarity-based methods consists of computing a similarity measure for each pair of items. This measure can be then used to cluster items, which is naturally interpreted as associating the items of a cluster with a single KC. In \cite{Rihak2017}, different measures of item similarity (Pearson, Cohen's Kappa, and Yule) were evaluated on real and simulated data. A different method for identifying the similarity between pairs of items, which is based on Fisher\lq{s} Exact Test of independence, was proposed in \cite{trevor_similarity} and was applied to data from an Introductory Physics MOOC. In addition to correct/incorrect information, `item-similarity' can be based also on other behavioral characteristics, such as response-times \cite{boros_response_times,Rihak2017}.

The item-similarity methods used in educational data mining for clustering items make the implicit assumption that the latent trait (mastery of the specific skill) is \textit{fixed} during the learning activity that generated the responses (so students' responses to items that belong to the same KC should be highly correlated). This assumption may be reasonable in the context of  \textit{testing} (summative assessment), which is expected to occur \textit{after} the learning process (in \cite{Rihak2017}, the authors explicitly refer to this shortcoming of the item-similarity measures, and mention that by using these methods ``we mostly ignore the issue of learning'', p. 17). 
However, \textbf{this assumption does not hold in the context of learning}. In such cases, the correlation between the items might not be a good indication of their similarity (e.g., students will tend to fail on the first items of each KC, and succeed on later ones). 

The goal of this research is to address this gap by providing a measure that can capture similarity in the context of \textit{learning}. For that, we propose a new item-similarity measure termed \textit{Kappa Learning} (KL).
The main assumption behind KL is that students' performance on items belonging to the same KC can be increasing, but not decreasing. As we use dichotomous scoring (correct/incorrect on first attempt), we expect that the performance of student $s$ on KC $k$ would take the form of a `step' function, which moves from 0 to 1 when $s$ masters $k$ (\textit{guess} or \textit{slip} may occur, and introduce noise). To quantify that, KL extends the notion of `agreement' in \textit{Cohen's Kappa}.

We first make the assumption that the items are administered to the students in the same order (defined by the instructional designers), but we later explain how our formula naturally generalizes to random or adaptive ordering. We note that we do not assume that all items belonging to the same KC will be presented to students one after the other, or that all the students attempt all the items. On the contrary, we assume that students can skip items, and that items from different KCs may interleave (as in the data that we analyze), which makes the clustering non-trivial. We then compare a clustering that is based on KL to clustering that is based on the similarity measures evaluated in \cite{Rihak2017}, and show that KL significantly outperforms them. 

The rest of this paper is organized as follows. In Section~\ref{KL}, we present Cohen's Kappa and our new measure, Kappa Learning. Section~\ref{met} describes the clustering method. The details of the empirical setting and the data are provided in Section~\ref{empirical}. In Sections~\ref{results} and~\ref{simulation} we evaluate the performance of Kappa Learning against standard similarity measures on real and simulated data, respectively. Finally, in Section~\ref{discussion} we discuss the results and suggest directions for future research. 






\section{Cohen's Kappa and Kappa Learning}
\label{KL}
\subsection{Cohen's Kappa}
\label{KM}
Cohen's Kappa (sometimes abbreviated as Kappa) is a measure of inter-rater agreement for nominal scales \cite{cohen1960coefficient}. 
\begin{equation}
\label{EQ:SK}
S_k = \frac{P_o - P_e}{1 - P_e}  
\end{equation}
where: \newline
$P_o$ is an \textit{observed} level of agreement \newline
$P_e$ is an \textit{expected} level of agreement \newline

The observed level of agreement is the proportion of the cases the raters agree upon. The expected level of agreement is the proportion of agreement that is expected by chance. 

We consider items as raters, learners as subjects to classify, and learners' responses as classification results. We interpret learner's correct/incorrect answer to an item (encoded as $1/0$) as the rater's (=item) attempt to identify if the learner has mastered the KC underlying the item. Let us consider a contingency table summarizing learners' responses to two different items: $Q_1$ and $Q_2$. Let us assume $n$ learners answered both items, and the number of learners in each cell is defined as follows (Table~\ref{tab:CT}): \newline
\begin{itemize}
\item $a$ - number of learners answered both $Q_1$ and $Q_2$ correctly
\item $b$ - number of learners answered $Q_1$ incorrectly and $Q_2$ correctly
\item $c$ - number of learners answered $Q_1$ correctly and $Q_2$ incorrectly
\item $d$ - number of learners answered both $Q_1$ and $Q_2$ incorrectly
\item $n$ - total number of learners ($n=a+b+c+d$)
\end{itemize}

\begin{table}[h!]
\centering
\caption{\label{tab:CT}A contingency table for $Q_1$ and $Q_2$.}
\begin{tabular}{cccc}
 & $Q_1$ correct & $Q_1$ incorrect\\\hline
$Q_2$ correct & a & b & a + b\\
$Q_2$ incorrect & c & d & c + d \\\hline
 & a + c & b + d & n\\
\end{tabular}

\end{table}

The number of cases the raters agree upon (the learner gave the same answer to both items) is equivalent to $a+d$. Intuitively, if two different items belong to the same skill, and learner's mastery of that skill is \textit{fixed} during the learning activity, the learner is expected to answer both items either correctly or  incorrectly, depending on whether the skill is mastered or not. So, it follows that:
\[P_o = \frac{a + d}{n} \]
The items are independent in the sense that each item independently `rates' if a learner belongs to a category of learners knowing a particular KC. So we could compute the level of agreement that is expected by chance as a sum of products of marginal probabilities (Table~\ref{tab:CT}). 
\[P_e = \frac{(a + b)(a+c) + (b+d)(c+d)}{n^2} \]

By doing substitution of $P_o$ and $P_e$ into Equation ~\ref{EQ:SK} and some straight forward simplification we get:
\[S_k = \frac{2(ad-bc)}{(a+b)(b+d) + (a+c)(c+d)} \]

\subsection{Kappa Learning: Adjusting Kappa to Accommodate Learning}
\label{KLM}
To accommodate for learning, we give a different interpretation to the notion of `agreement' in Cohen's Kappa formula, taking into account possible improvement of learner's skill, or in other words, learning. \newline

\noindent We make the following assumptions on the process:
\begin{enumerate}
\item The items are presented to the learners in a fixed order\footnote{We later remove this assumption}.
\item The items belong to k>1 KCs; Each item belongs to one KC; Items belonging to different KCs may interleave (which makes the clustering non-trivial)
\item Learner's success on items belonging to the same KC behaves like a `step' function: Before mastering the skill of $KC_m$, the student fails on items of  $KC_m$; once mastering the skill underlying  $KC_m$, the student succeeds on future items of $KC_m$ (\textit{guess} and \textit{slip} may occur; we assume no `forgetting'). 
\end{enumerate}

For a pair of items $Q_1$, $Q_2$, where $Q_1$ is presented to the learners \textit{before} $Q_2$, we define the values in the contingency table (Table~\ref{tab:CT}) as follows:
\begin{itemize}
\item $a$ - number of learners who got both items correct, namely mastered the required skills before getting to $Q_1$. This is a case of agreement. 
\item $b$ - number of learners who got the first item incorrect and the second item correct, namely, mastered the required skill after getting to $Q_1$, but before getting to $Q_2$. \textbf{This is a case of agreement, and is where our measure differs from Cohen's Kappa}
\item $c$ - number of learners who got the first item correct and the second item incorrect. This is the only case interpreted as \textit{disagreement}. 
\item $d$ - number of learners answered both $Q_1$ and $Q_2$ incorrectly. This is a case of agreement.
\item $n$ - total number of learners ($n=a+b+c+d$)
\end{itemize}

Based on these, we define $P_o$ and $P_e$ as follows:
\[P_o = \frac{a + b + d}{n} \]
\[P_e = \frac{(a+b)(b+d) + (a + b)(a+c) + (b+d)(c+d)}{n^2} \]

By doing substitution of $P_o$ and $P_e$ into Equation ~\ref{EQ:SK} and some straight forward simplification we get:
\begin{equation}
S_{kl} = \frac{(ad-bc)}{(a+c)(c+d)}
\end{equation}
We call this measure Kappa Learning and denote it  $S_{kl}$. The values of both Kappa and Kappa Learning range between $-1$ and $+1$, where $0$ means independence, and $+1$ means perfect agreement. In Kappa it is achieved when both $c$ and $b$ are equal to $0$. In Kappa Learning, perfect agreement is achieved when $c$ equals $0$.

\subsection{Generalizing to Random Order of Items}
\label{ROQ}
We now explain how the assumption that items are presented in fixed order can be removed. Let us assume that the items administered to the learners in a random order, meaning that different learners may see the items in different order.  In this case, for each learner and for each pair of items we construct the contingency table (similar to Table~\ref{tab:CT}) by computing the values of a, b, c, d as follows:
\begin{itemize}
\item $a$ - number of learners who got both items correct
\item $b$ - number of learners who got the first item presented to them (among $Q_1$ and $Q_2$) correct, and the second incorrect
\item $c$ - number of learners who got the first item presented to them (among $Q_1$ and $Q_2$) incorrect, and the second correct
\item $d$ - number of learners who got both items incorrect
\end{itemize}

\section{Method}
\label{met}

\subsection {Similarity Measures}
To evaluate the performance of Kappa Learning (denoted $S_{kl}$), we compare it to the following similarity measures:

\begin{itemize}
\item $S_k$: Cohen's Kappa inter-rater agreement
\item $S_p$: Pearson product-moment correlation coefficient
\item $S_y$: Yule coefficient of association
\end{itemize}

Cohen's Kappa (see also in Subsection~\ref{KM}) coefficient is defined as:
\begin{equation}
S_k = \frac{2(ad-bc)}{(a+b)(b+d) + (a+c)(c+d)} 
\end{equation}

Pearson product-moment coefficient is a measure of linear  correlation between two variables. When applied to dichotomous data, the Pearson correlation coefficient returns the phi ($\phi$) coefficient. So, in terms of $a$, $b$, $c$ and $d$ (Table~\ref{tab:CT}) the value of $S_p$ is computed as follows:
\begin{equation}
S_{p} = \frac{(ad-bc)} {\sqrt[]{(a+c)(a+b)(b+d)(c+d)}}
\end{equation}

Yule coefficient of association is a measure of colligation between two binary variables and it is commonly used for analyzing scores in Item Response Theory (IRT). It is the number of pairs in agreement ($ad$) minus the number in disagreement ($cb$) over the total number of paired observations and it is defined as: 
\begin{equation}
S_{y} = \frac{(ad-bc)}{(ad + bc)}
\end{equation}

All three measures range from minus unity to unity, where $1$ indicates perfect agreement, $-1$ indicates perfect disagreement, and $0$ indicates no relationship \cite{warrens2008association}.
A thorough evaluation of these measures as means for clustering items in an interactive learning environment was done by  {\v{R}}ih{\'a}k and Pel{\'a}nek \cite{Rihak2017} (they analyzed the most appropriate measures among the $76$ measures analyzed in \cite{Choi10asurvey}. Jaccard and Sokal, which were also evaluated in \cite{Rihak2017}, produce much lower results on our data and are therefore omitted from the analysis).

We follow a similar methodology to the one proposed in \cite{Rihak2017}, described below, and demonstrate that Kappa Learning outperforms the other measures.

\subsection{Process}
Our process has two main steps: 1) Cluster the items based on the four similarity measures (Kappa Learning, and the three reference measures). 2) Compare the goodness-of-fit of the clusterings computed in step 1.

\noindent \textbf{Step 1.} Computing the clustering includes the following sub-steps (per similarity measure):

\begin{enumerate}

\item Using students' performance data we first compute \textit{user-based} item similarity matrix, denoted $M1$. $M1[i,j]$ contains the result of the relevant similarity measure for items $q_i$ and $q_j$. 

\item Compute \textit{item-based} distance matrix from the user-based similarity matrix $M1$. The rationale is that for a pair $(q_i, q_j)$, if $q_i$ and $q_j$ are similar (i.e., belong to the same KC), they should have a similar distance to a third item $q_k$ (whether it is in the same KC or not). This incorporate more information into the similarity between the items, which should improve the accuracy of the clustering \cite{Rihak2017}. We denote the \textit{item-based} distance matrix with $M2$. Two standard metrics are used for computing $M2$: Pearson and Euclidean.


\item  Two clustering algorithms  are applied on $M2$: K-Means and Ward's Hierarchical \cite{jain2010data}. The number of clusters is derived from the hierarchal Knowledge Tree defined by content experts (see Subsection~\ref{KT}). 

\end{enumerate}

\noindent \textbf{Step 2.} Per clustering, we use Adjusted Rand Index (ARI) \cite{hubert1985comparing,10.2307/2284239} to measure the goodness-of-fit against ground truth -- experts' mapping of the items into the knowledge tree.  

ARI is a common measure for comparing the similarity between two clusterings. In ARI, similarity is interpreted as the amount of pairs of items on which the clusterings `agree', adjusted for the amount of agreement `by chance'. 

To be concrete, assume $C$ is a dataset which contains $m$ items, with two clusterings of $C$ into $k$ clusters, denoted $C_1$ and $C_2$. For a pair of items $(i_1, i_2)$, $C_1$ and $C_2$ `agree' on $(i_1, i_2)$ iff $i_1$ and $i_2$ are either assigned to the same cluster, or to different clusters, in both $C_1$ and $C_2$. 

To evaluate the level of agreement between $C_1$ and $C_2$, we define a contingency table with the values $a$, $b$, $c$, and $d$, as follows:

\begin{itemize}
\item $a$ - number of pairs $(i_1, i_2)$ where $i_1$ and $i_2$ are assigned to the \textit{same} cluster in $C_1$ and in $C_2$. This is a case of agreement. 
\item $b$ - number of pairs $(i_1, i_2)$ where $i_1$ and $i_2$ are assigned to the \textit{same} cluster in $C_1$ and to \textit{different} clusters in $C_2$. This is a case of disagreement. 
\item $c$ - number of pairs $(i_1, i_2)$ where $i_1$ and $i_2$ are assigned to \textit{different} clusters in $C_1$ and to the \textit{same} cluster in $C_2$. This is a case of disagreement.  
\item $d$ - number of pairs $(i_1, i_2)$ where $i_1$ and $i_2$ are assigned to \textit{different} cluster in $C_1$ and in $C_2$. This is a case of agreement.  
\end{itemize}

The total number of pairs, $n$, is equal to $a+b+c+d = \frac{m(m-1)}{2}$.

Using this definition of $a$, $b$, $c$, and $d$, we can construct a contingency table similar to Table~\ref{tab:CT} for pairs of items, and compute Cohen's Kappa based on this table, which is equivalent to Adjusted Rand Index \cite{warrens2008equivalence}. 

\section{Empirical Settings}
\label{empirical}
\subsection{The Learning Environment}
\label{LE}
We use data from a computerized tutor that teaches Fractions for \nth{4} grade. The students progress through the Tutor on their own pace, in a linear order defined by the content experts. The subject matter knowledge that the Tutor covers is  modeled by a Knowledge Graph, which is described in Subsection~\ref{KT} (since it is hierarchical, hereafter we use the term Knowledge Tree). 

The content of the Tutor includes $550$ items, instructional materials such as videos, and on-line labs that students can use to explore the various concepts. These are arranged in $112$ learning units. Each of the learning units contains a collection of items and learning materials, and is designed to take approximately $5-15$ minutes.

The course is divided into two parts. Part A contains $57$ learning units which include $337$ items, and Part B contains $55$ learning units which include $213$ items. Concepts are first introduced and explained, and are later re-visited. This means that items which require a certain skill can appear on different locations.

\subsection{Knowledge Tree and Content Mapping}
\label{KT}
The course was designed according to a Knowledge Tree (KT) that models the hierarchy of skills that students should master (under the root topic ``Fractions for \nth{4} grade''). The KT was developed by the content experts who built the course. The first level, termed `subject', includes $8$ topics. Some of these subjects have second level, termed `sub-subject'. On this level of the tree (sub-subject + subjects that do not have second level) there are $19$ topics. The division of the first two levels is curricular (e.g., 'adding fractions' as first level, with `adding fractions with a common denominator' and `adding fractions with a different denominator' as its children). 

\begin{figure}[h!]
\label{Diagram of Knowledge Components}
\centering
\includegraphics[width=0.5\textwidth]{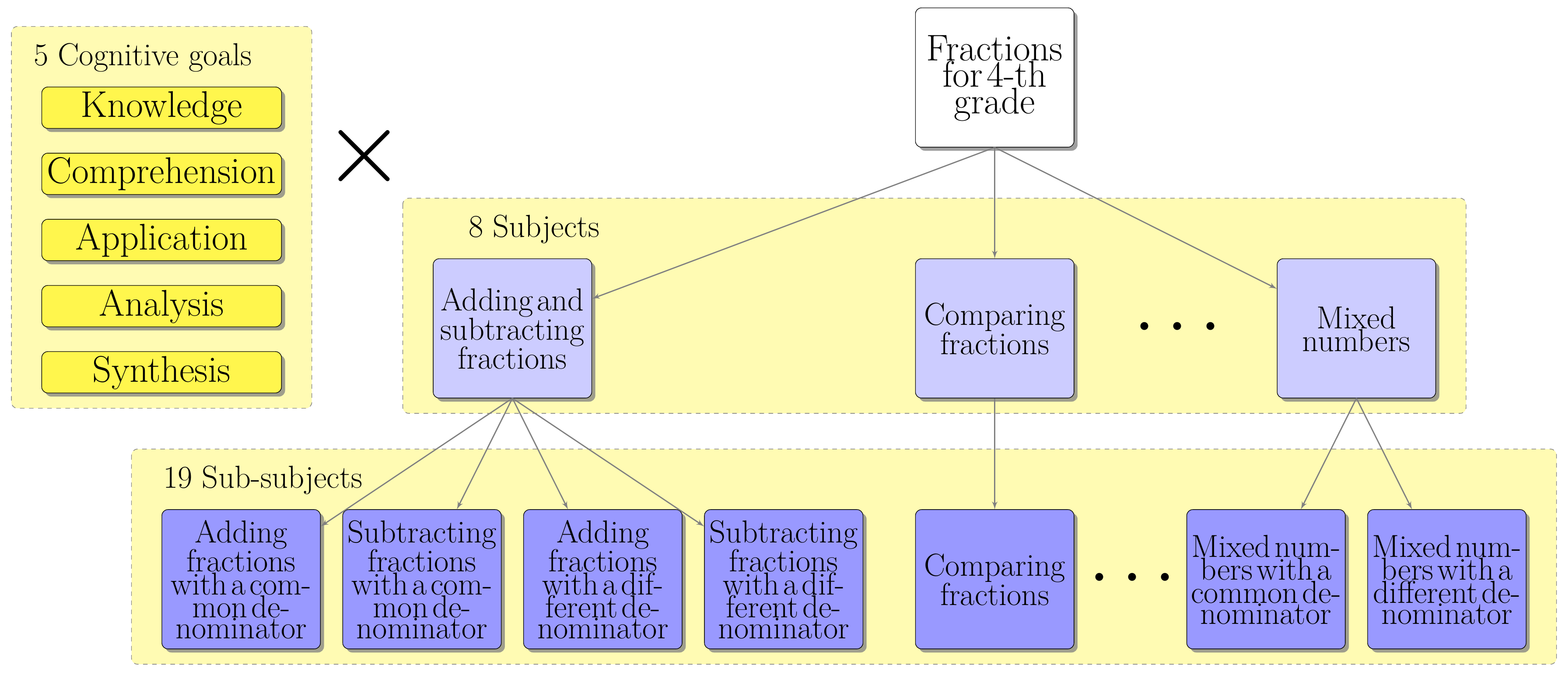}
\caption{\label{fig:SKT} Content expert's Knowledge tree for the topic ``Fractions for  \nth{4} grade''. The refining of subjects/sub-subjects into 'first level + goal'/'second level + goal' is computed as a Cartesian product of goals layer with subjects/sub-subjects layers correspondingly.}
\end{figure}

In addition to these two levels, there is a third level, termed `goals`, which is orthogonal to the classification into subject/sub-subject, and refers to the cognitive type of the task (it resembles Bloom's Taxonomy \cite{anderson2001taxonomy}). Since the `goal' level is orthogonal to the division into subjects/sub-subjects (Figure~\ref{fig:SKT}), it can be interpreted as refining the categories under `subject' (1st level + goals), or as refining the `sub-subject' (second level + goals).

The experts tagged each item with the `subject', `sub-subject', and `goal' it belongs to. On most cases, each item is mapped  into one category on each level. In the few cases were an item was mapped into more than one category, we assume that each unique combination of subjects/sub-subjects is actually a new knowledge concept (similar to the rationale of \cite{peter_subskills}). 
For example,  if item $i$ is marked as belonging to subjects 1 and 2, we create a new artificial subject for this combination of subjects. We removed from the data artificial combinations containing only one item, and the few items ($<5$) that belong to these combinations.  

\subsection{Knowledge Components}
We interpret Knowledge Component (KC) as a group of items that deal with the same concept (i.e., require the same skill) \footnote{We use the term KC in two ways -- as \textit{skill}, and as a \textit{set of items} that require a certain skill}. We examine classifications of items into Knowledge Components that are based on different levels of granularity with respect to the Knowledge Tree. For example, `First level' is a classification that is based only on the first split of the tree (`subject'). Table~\ref{tab:Clusters} presents the number of KCs defined by each level of the KT. 

\begin{table}[h!]
\centering
\caption{\label{tab:Clusters} Number of Knowledge Components by the level of Knowledge Tree.}
\begin{tabular}{cc}
Level of & Number of\\
Knowledge Tree & Knowledge Components \\
\hline
First&14\\
First with Goals&42\\
Second& 32\\
Second with Goals&62\\
\end{tabular}
\end{table}

\subsection{Data}
The data include the responses of $594$ \nth{4} grade students, who used the Tutor for a few hours a week during regular class hours, for a period of $2$ months. (We remove the data of students who attempted less than $50$ items, and the few who had less than $25\%$ success on first attempt, as we assume they were mainly `gaming the system'). On average, students spent about $12$ hours in the Tutor. 

Students' performance is operationalized as \textit{correct on first attempt}. From the log files, we build a $0/1$ $student \times item$ response matrix, denoted $RM$. $RM[i,j]$==1 $iff$ students $i$ solved item $j$ correctly on first attempt. 

\section{Results on Real Data}
\label{results}
\subsection{Computing the Similarity Matrix}
We compute similarity matrix for each of the four measures, as described in Section~\ref{met}. This yields four similarity matrices.

\noindent To cluster the items based on these matrices, we use three clustering algorithms: 
\begin{itemize}
\item Ward's Hierarchical clustering using Pearson correlation Distance
\item Ward's Hierarchical clustering using Euclidean Distance 
\item K-means clustering using Euclidean Distance 
\end{itemize}

As noted before, the number of clusters is defined according to the number of Knowledge Components of the Knowledge Tree (Table~\ref{tab:Clusters}). Goodness-of-fit of a clustering is evaluated by measuring its similarity to the ground truth labeling, using Adjusted Rand Index (ARI). 

\subsection{Results of Hierarchical Clustering}
Table~\ref{tab:HC} 
demonstrates the results of the Hierarchical Clustering on the entire course, based on the four similarity measures, using Pearson Distance (which outperforms Euclidean Distance in all combinations; thus we omit the results for Euclidean Distance). As can be seen, clustering that is based on Kappa Learning outperforms the other measures in all the combinations. 

\begin{table}[h!]
\centering
\caption{\label{tab:HC} Adjusted Rand Index for different similarity measures, using Hierarchical Clustering and Pearson Distance, for number of KCs that is based on different levels of the Knowledge Tree.}
\begin{tabular}{ccccc}
\hline
& & First & & Second\\
&  First &  with Goals & Second & with Goals\\\hline
\textbf{Kappa Learning} & \textbf{0.26} & \textbf{0.21} & \textbf{0.26} & \textbf{0.36}\\
Kappa & 0.16& 0.17& 0.18& 0.27\\
Yule & 0.15& 0.19& 0.21& 0.29\\
Pearson& 0.16& 0.18& 0.21& 0.30\\\hline
\end{tabular}
\end{table}

\subsection{Results of K-Means Clustering}
\begin{figure}[h!]
\label{K-Means_results}
\centering
\includegraphics[width=0.5\textwidth]{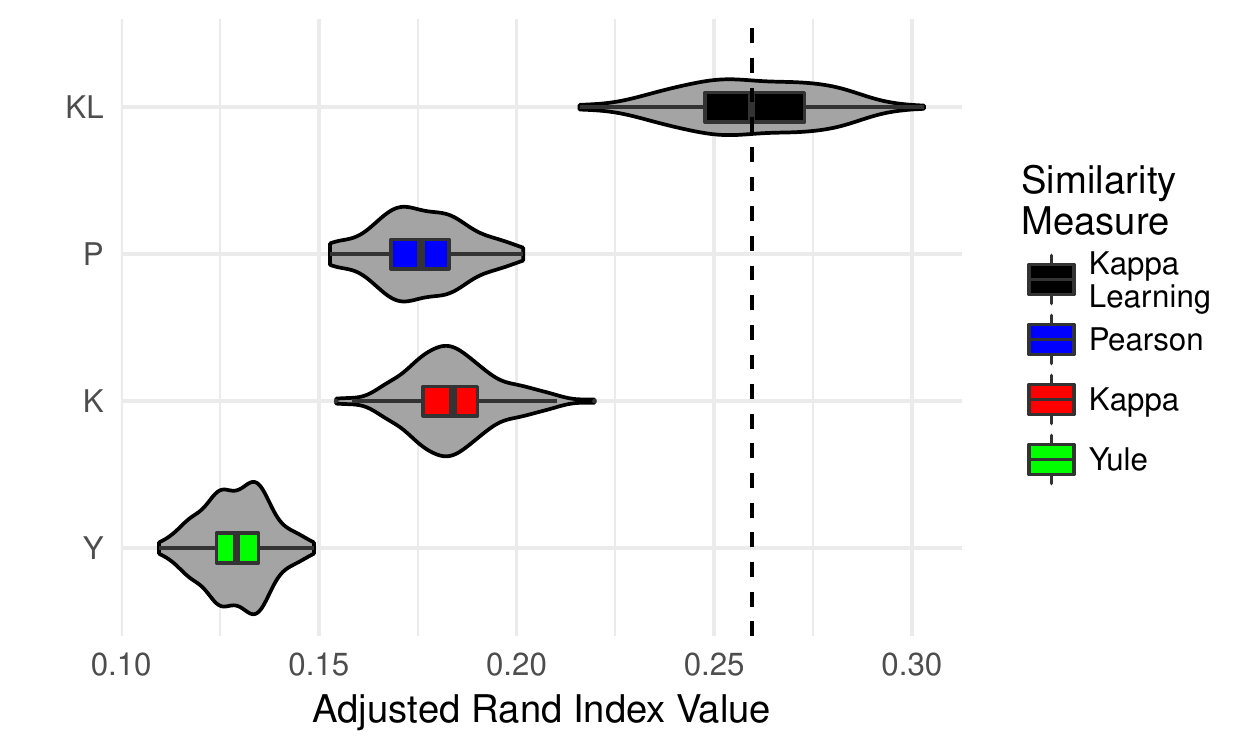}
\caption{\label{fig:2KC}Results of K-means clustering for the entire course and number of KCs defined by the second level of the Knowledge Tree. The vertical dashed line goes through the mean of the distribution of the ARI results for Kappa Learning.}
\end{figure}

\begin{figure}[h!]
\centering
\includegraphics[width=0.5\textwidth]{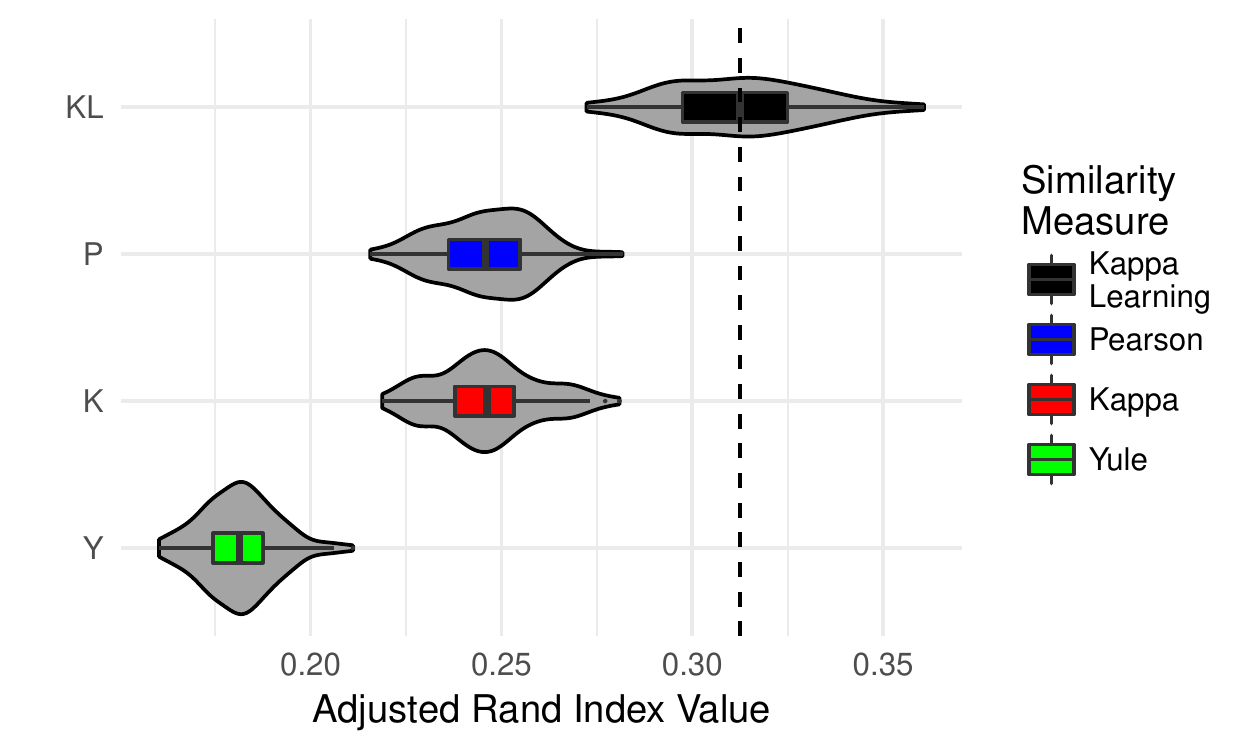}
\caption{\label{fig:2KCG} Results of K-means clustering for the entire course and number of KCs defined by the second level of the Knowledge Tree $+$ Goals. The vertical dashed line goes through the mean of the distribution of the ARI results for Kappa Learning.}
\end{figure}

In addition to the comparison that is based on Hierarchical Clustering, we make a comparison that is based on K-Means Clustering. Since K-Means is non-deterministic (depends on random assignment of initial cluster centers), we run the algorithm $100$ times for each combination, each time computing the Adjusted Rand Index against ground truth. The distribution of the Adjusted Rand Index for each combination are presented in Figures~\ref{fig:2KC} and~\ref{fig:2KCG}. As can be seen, Kappa Learning outperforms the other similarity measures. 


\subsection{Finding optimal number of clusters}
Finding an optimal number of clusters is a fundamental problem in clustering analysis that is typically ill-posed \cite{hubert1985comparing}, as there is no rigorous definition of a cluster, and the practical considerations are domain and application-specific. For example, in our model we consider number of clusters that is based on different resolution of experts' hierarchical Knowledge Graph  (Subsection~\ref{KT}). 

The `goodness' of the resulting clustering is usually measured by cluster cohesion and cluster separation. One of the measures for cluster cohesion or compactness is Within Cluster Sum of Squares (WSS), $W_k$. For any clustering of a set $S$ into $k$ clusters $S = \{S_1, S_2, \ldots ,S_k\}$, WSS is defined as 
\begin{equation}
W_k = \sum\limits_{i=1}^k \frac{1}{2|S_i|}\sum\limits_{x,y \in S_i} [dist(x,y)]^2
\end{equation}
where $dist(x,y)$ is a measure for distance between two items of a set. 

In our case the value of $W_k$ depends on the method used for evaluating the item's similarity matrix based on student's performance matrix, the method for measuring the distance between items of similarity matrix and the clustering algorithm used. Within Clusters Sum of Squares is commonly used to find an optimal number of clusters using the `elbow' heuristic, however in our case there is no clear `elbow' in the graph. Another common method for estimating an optimal number of clusters using WSS measure is the Gap statistic method. 

\subsubsection{Gap Statistic}
The main idea of Gap statistic is comparing the goodness of clustering applied to a specific dataset with the goodness of clustering obtained when applied on a uniformly distributed data with no clustering structure at all (so called 1-cluster data) \cite{tibshirani2001estimating}. The $GAP_k$ measure used in Gap statistic is the difference between an expected value of $\log(W_k)$ computed for clustering of 1-cluster random data into $k$ clusters and $\log(W_k)$ value obtained from clustering of input dataset into the same number of clusters $k$. The random data is generated from a uniform distribution over the same range as the input dataset. The Gap statistic method receives $K.max$ -- the maximal number of clusters to consider, a clustering algorithm, a distance measure, and an input dataset. For each $k$ from $1$ to $K.max$, it computes $GAP_k$ value and searches for the value of $k$ that maximizes the Gap value.

For the four similarity matrices obtained (Section~\ref{met}) we compute Gap statistic using   Ward's Hierarchical Clustering, Pearson distance and $K.max = 70$. Then we apply two different methods for computing the optimal number of clusters: First SE Max (first local maximum of Gap value within one standard error) and First Max (first local maximum of Gap value) \cite{tibshirani2001estimating}. 

\begin{figure}[h!]
\centering
\includegraphics[width=0.5\textwidth]{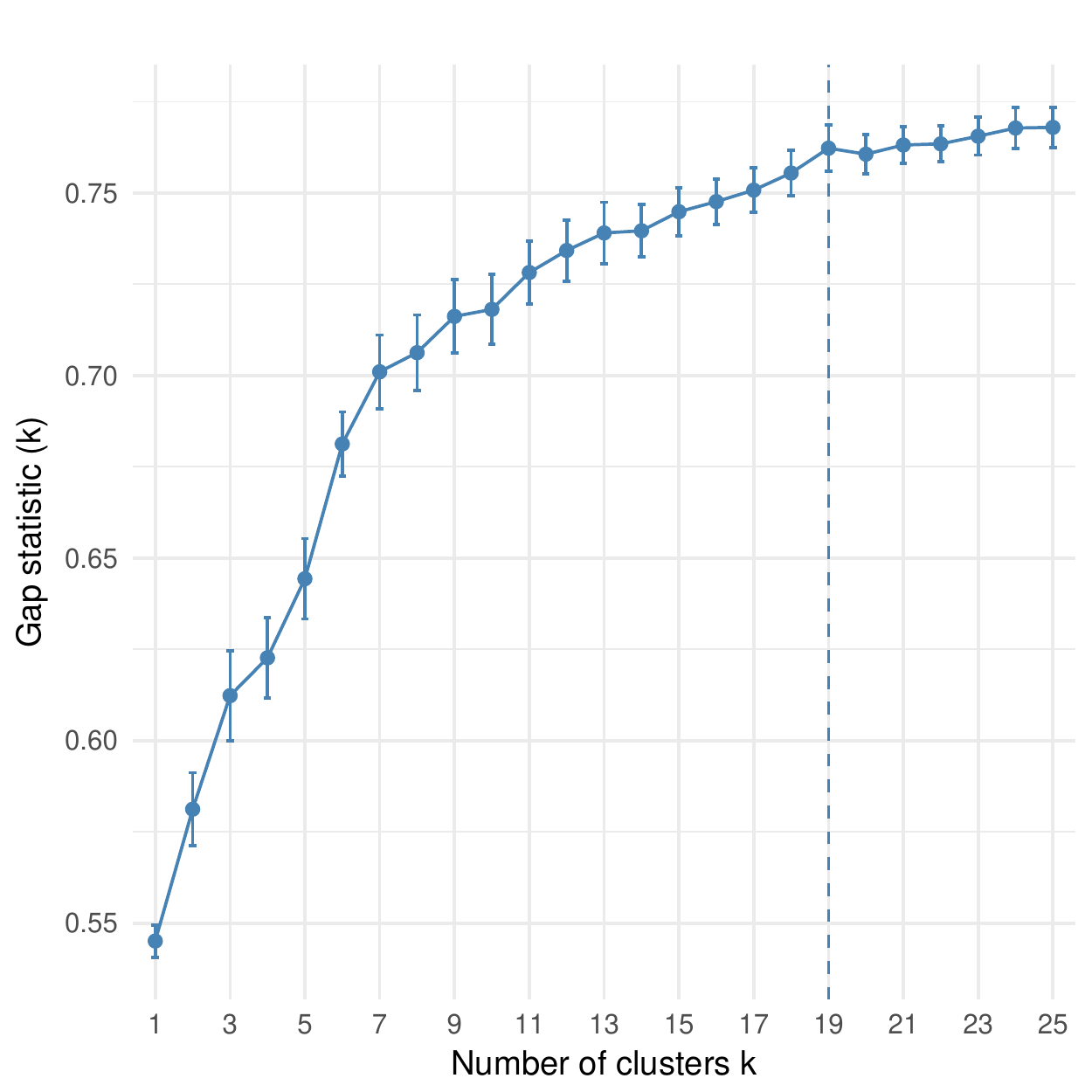}
\caption{\label{fig:GAP} The values of GAP statistic for $k$ in a range from $1$ to $25$ computed based on Kappa Learning similarity matrix. The vertical dashed line indicates the optimal number of clusters as predicted by both firstMax and firstSEMax methods.}
\end{figure}

Running Gap statistic on Kappa Learning dataset produces $19$  as an optimal number of clusters (Figure~\ref{fig:GAP}), which is similar to the number of Knowledge Components at the Second level of the Knowledge Tree (Figure~\ref{fig:SKT}; the second level of the Knowledge Tree is denoted `sub-subjects'). 

The optimal number of clusters based on Yule similarity measure is $14$, which is also quite close to the ground truth. With the two other methods (Cohen's Kappa, Pearson), Gap statistic does not produce meaningful results (Table~\ref{tab:GAP}).  

\begin{table}[h!]
\centering
\caption{\label{tab:GAP} Optimal number of clusters by GAP statistic.}
\begin{tabular}{ccc}
\hline
& First Max& First SE Max\\
& Method & Method\\\hline
\textbf{Kappa Learning} & \textbf{19} & \textbf{19} \\
Kappa & 1 & 1\\
Yule & 15 & 14\\
Pearson & 1 & 1\\\hline
\end{tabular}
\end{table}

\section{Simulation Study}
\label{simulation}
In addition to evaluating our method on data from a real learning environment (Subsection~\ref{LE}),  we conduct a simulation study. 

\subsection{Data Generation}
Our simulation model makes the following assumptions: 
\begin{itemize}
\item Each item belongs to one of $K$ knowledge components (KCs); the items are uniformly distributed among these KCs. Each KC has an individual difficulty level (drawn from a probability distribution defined below). 
\item The order of appearance of KCs is predefined. We assume that the topic is first presented and explained to the learners, so the majority ($\approx 60\%$, chosen empirically based on the data) of the items that belong to it appear one after the other. The rest of the items that belong to the KC are presented to the learner on a later stage, and interleaved between items from other KCs. 
\item Students learn as they interact with the items; Learners have individual learning rate (drawn from a probability distribution defined below).
\end{itemize}
\textit{}

\subsubsection{Hidden Markov Model and Bayesian Knowledge Tracing}
Bayesian Knowledge Tracing (BKT) \cite{corbett1994knowledge} is a popular approach to model skill acquisition in Intelligent Tutoring Systems. It models a student knowledge as a latent binary variable of a Hidden Markov Model. Learning is modeled as a transition from `not mastered' to `mastered' state. The standard BKT model uses the same four parameters for all the students and items. Several studies extended the basic BKT model with individualized parameters for student ability and item difficulty (e.g., \cite{pardos2011kt, khajah2014integrating, yudelson2013individualized}). We  use the model introduced in \cite{yudelson2013individualized} as the underlying model for the data generation process. 

\subsubsection{Individualized Bayesian Knowledge Tracing}
We apply Individualized BKT approach with parameter splitting \cite{yudelson2013individualized} to model a learning process. Namely, we construct individual HMM per student and KC. All items of the same KC are assumed to have the same difficulty. The model assumes students learn as they practice more. On each opportunity to solve an item that belongs to a knowledge concept, the probability that the student masters the skill underlying the item's KC increases. 

Let us define:
\begin{itemize}
\item $L$ - number of learners 
\item $K$ - number of Knowledge Components
\item $N$ - total number items (questions)
\end{itemize}

For each KC $k$ and each student $l$ we generate the following parameters:
\begin{itemize}
\item $P(L_0)$ - the probability that a student initially knows a particular KC. In this model we assume the students have no initial knowledge.
\item $P(T)_l^k$ - the probability of learning for student $l$ and skill $k$. 
\item $P(S)$ - probability of slip, meaning making an incorrect attempt when applying a known skill. We assume $P(S) = 0.1$ (not individualized).
\item $P(G)$ - probability of random guess, meaning making a correct attempt when applying unknown skill. We assume $P(G) = 0.2$ (not individualized).
\end{itemize}

As proposed in \cite{yudelson2013individualized}, the value of the parameters $P(T)_l^k$ is combined from two components: a per-skill component and a per-student component. So, we generate a pair $(P(T)_l,P(T)^k)$ of parameters for each skill and each student. Each of the above parameters is generated from a uniform distribution $\mathcal{U}(0,1)$. Then for each student $l$ and KC $k$ the parameters are combined as follows:
\begin{equation}
P(T)_l^k = \sigma (l(P(T)_l) + l(P(T)^k))
\end{equation}
where:
\begin{equation}
\sigma (x) = 1/(1+exp^{-x}) 
\end{equation}
and 
\begin{equation}
l(x) = \log{(x/1-x}) 
\end{equation}

Where $\sigma (x)$ and $l(x)$ are the sigmoid and logit functions, respectively. 

The performance matrix for each student and knowledge concept is generated using R's HMM package, and the data is combined into a $L \times N$ student's performance matrix. For all Knowledge Components containing more than $6$ items, the first $6$ items are placed one after another, modeling the introduction of the concept to the learners. The rest of the items are shuffled randomly between future KCs.

\subsubsection{Model Parameters}
In the experiment reported below the basic setting is $1000$ learners, $20$ Knowledge Components, $200$ items. The parameters are chosen in a way that approximates the multivariate distribution of the real data with respect to the average number of items per Knowledge Component and the mean performance of students, as illustrated in Table~\ref{tab:Model}. 

\begin{table}[h!]
\centering
\caption{\label{tab:Model} Comparison of simulation model to empirical data.}
\begin{tabular}{ccc}
\hline
&  Average number & Average \\
&  of Questions  & performance  \\
&  per KC  & of Students \\\hline
Empirical data & $\approx $9 & 0.67\%\\
Simulation model & 10 & 0.64\%\\\hline
\end{tabular}
\end{table}

To evaluate the clustering that is based on each of the four measures, we follow the same process as described in Section~\ref{met}. Since the results depend on the simulated data, we repeat the process $700$ times, each time starting with generating a new performance matrix. 

\subsection{Results on Simulated Data}
\begin{figure}[h]
\centering
\includegraphics[width=\columnwidth]{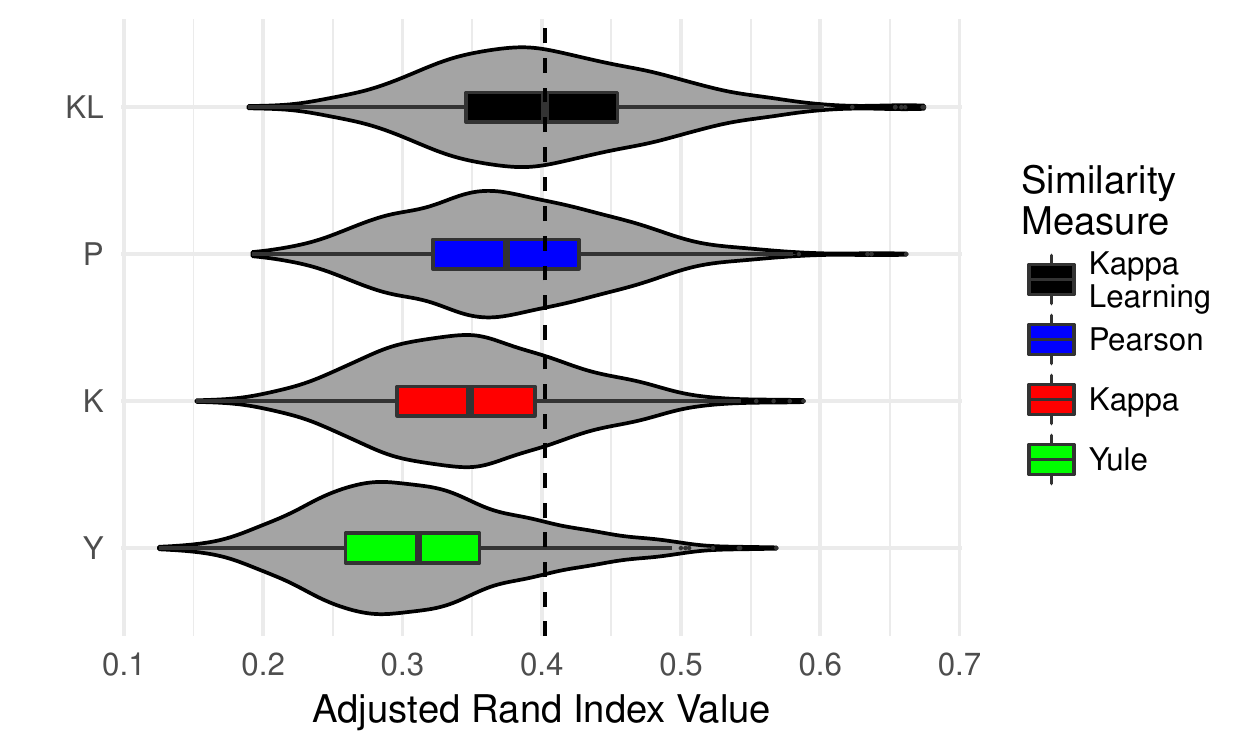}
\caption{\label{fig:SimARI} Distribution of Adjusted Rand Index values for different similarity measures (KL - Kappa Learning, K - Kappa, Y - Yule, P - Pearson). The vertical dashed line goes through the mean of the distribution of ARI values for Kappa Learning. }
\end{figure}

The results are presented in Table~\ref{tab:SimRes} (right column) and Figure~\ref{fig:SimARI}. As can be seen, Kappa Learning outperforms all other measures in its ability to reproduce the original clusters.
Table~\ref{tab:SimRes} also presents the results of each measure on the real data (left column), for reference.

\begin{table}[h!]
\centering
\caption{\label{tab:SimRes} Comparison of Adjusted Rand Index values for different similarity measures.}
\begin{tabular}{ccc}
\hline
& Real data,  &  Simulation\\
& Second level &  Model\\
&  with Goals  & (averaged over 700 runs)\\\hline
\textbf{Kappa Learning} &  \textbf{0.36} & \textbf{0.40}\\
Kappa & 0.27& 0.35\\
Yule & 0.29& 0.31\\
Pearson& 0.30& 0.37\\\hline
\end{tabular}
\end{table}

To verify the statistical significance of the results, we conduct a t-test for the results of Kappa Learning vs. the three other measures (Yule, Cohen's Kappa, and Pearson). For all combinations, the p-value is less than $0.01$.

\section{Discussion}
\label{discussion}
The results show that Kappa Learning - the new similarity measure that we propose, which is based on adjusting Cohen's Kappa to `learning', can improve the clustering of educational items into Knowledge Components, compared to the state-of-the-art (the measures that are reported in \cite{Rihak2017} as producing the best results). We ascribe this to the fact that Kappa Learning explicitly models similarity under the assumption that students' skill can grow during the activity (= learning), while the conventional measures are based on the assumption that students' skill is fixed.

On real data, with different combinations for the amount of clusters (Table~\ref{tab:Clusters}), the improvement with Hierarchical Clustering was in the range of $10-60\%$ (Table~\ref{tab:HC}), comparing to the conventional measures (Kappa, Yule, and Pearson). On simulated data that follow the `mastery' assumption, and allow items of different Knowledge Components to interleave (which makes the task more difficult; if all the items of a certain KC are presented together, the clustering is almost trivial), the improvement with Hierarchical Clustering was in the range of $10-20\%$ (Table~\ref{tab:SimRes}).

In real-life scenarios, the number of clusters, which the clustering algorithms that we use take as input, is typically unknown, and it is necessary to extract it from the data. On the task of finding an optimal number of clusters, Gap statistic on clustering that is based on Kappa Learning yielded a number of clusters ($19$) that is similar to the number of clusters in the ground truth (according to the second level of the Knowledge Graph. See Table~\ref{tab:Clusters}). Among the other measures, Gap statistic on Yule-based clustering also produced results that are reasonably close to the ground truth. For Kappa and Pearson, Gap statistic did not yield meaningful results.   

Overall, Kappa Learning was superior with respect to all the factors that were measured: Various interpretations of the ground truth (deciding on the KCs according to different levels of Knowledge Graph); clustering algorithm -- K-Means and Hierarchical Clustering; real and simulated data; and in reproducing the number of clusters with Gap statistic. Thus, we conclude that in the context of learning in structured  domains (such as K-6 Math), Kappa Learning provides a significant improvement to the task of clustering that is based on item similarity, compared to conventional item-similarity measures.


\subsection{Future Work}
This work provides a few directions for future research. On the next step, we intend to work with the developers of the computerized tutor on using the results of Kappa Learning to refine and optimize the pedagogic design (cognitive modeling, but also questions such as which KCs require more content, are too difficult, too easy, etc.). 

Algorithmic directions include studying additional ways to insert the notion of `learning' into existing item-to-skill detection methods, and additional sources of information such as domain experts or analysis of the body of the items (text, images, mathematical symbols, etc.).

In terms of use cases, it would be interesting to evaluate Kappa Learning on data from a variety of learning environments (e.g., MOOCs) and subject matters, and in particular, on domains in which knowledge is less structured (e.g., reading comprehension).

\subsection{Summary and Conclusions}
This paper presents a new method for measuring the similarity between educational items, termed Kappa Learning. The novelty of this method, compared to previous measures of similarity between educational items, lies in the fact that it explicitly captures the notion of `learning', namely, change of the latent trait (student's mastery of the concept). This is done by extending the notion of `agreement' within Cohen's Kappa basic formula. 

Our results show that clustering that is based on Kappa Learning outperforms clustering that is based on conventional methods (Cohen's Kappa, Yule, Pearson), on real data from K-6 Math Tutor that teaches multiple concepts, and on generated data that simulates learning of multiple, interleaved concepts.  
Thus, we believe that Kappa Learning is more suitable than existing measures for computing similarity between items in the context of learning in structured domains. 


\begin{acks}
The authors would like to thank The Center for Educational Technology for giving us access to the data, and to Hagar Rubinek and Yael Weisblatt for their help in conducting this research. This  research was supported by The Willner Family Leadership Institute for the Weizmann Institute of Science, Iancovici-Fallmann Memorial Fund, established by Ruth and Henry Yancovich, and by Ullmann Family Foundation. The authors would like to thank Ido Roll for his useful comments.
\end{acks}

\bibliographystyle{ACM-Reference-Format}
\bibliography{KL-bibliography}

\end{document}